\pdfoutput=1
\PassOptionsToPackage{table,dvipsnames}{xcolor}
\documentclass[11pt]{article}

\usepackage[final]{acl}

\usepackage{times}
\usepackage{latexsym}

\usepackage[T1]{fontenc}

\usepackage[utf8]{inputenc}

\usepackage{microtype}
\usepackage{amsmath,amssymb,amsfonts}
\usepackage{algcompatible}
\usepackage{algorithm}
\usepackage{algpseudocode}
\usepackage{cite}
\usepackage{hyperref}

\usepackage{inconsolata}

\usepackage{graphicx}
\usepackage{textcomp}
\usepackage{xcolor}
\def\BibTeX{{\rm B\kern-.05em{\sc i\kern-.025em b}\kern-.08em
    T\kern-.1667em\lower.7ex\hbox{E}\kern-.125emX}}

\usepackage{multirow}
\usepackage{array}
\usepackage{nicematrix}
\usepackage{booktabs}
\newcommand{\tabitem}{~~\llap{\textbullet}~~}
\usepackage{enumitem}
\usepackage{tabularx}
\usepackage{xspace}
\usepackage{amsmath, amssymb}
\usepackage{arydshln}
\usepackage{inconsolata}
\usepackage{soul}
\usepackage{makecell}
\usepackage{listings}
\usepackage{caption,subcaption}
\usepackage{soul}
\usepackage{anyfontsize}
\usepackage{tcolorbox}
\usepackage[author={An}]{pdfcomment}

\newcommand{\system}[1]{\textsc{#1}\xspace}  
\newcommand{\QQFS}{\system{QQSUM}}  

\newcommand{\QQSUMM}{\system{QQSUM-RAG}}

\newcommand{\dataset}[1]{\textsc{#1}}
\newcommand{\AmazonKP}{\dataset{AmazonKP}\xspace}

\newcolumntype{?}{!{\vrule width 1pt}}

\graphicspath{ {./images/} }

\makeatletter
\def\blfootnote{\gdef\@thefnmark{}\@footnotetext}
\makeatother

\definecolor{Cloud}{rgb}{0.949,0.953,0.960}
\definecolor{Oat}{rgb}{0.910,0.886,0.784}
\definecolor{Sunrise}{rgb}{1,0.941,0.941}

\newcommand{\ctext}[3][RGB]{%
  \begingroup
  \definecolor{hlcolor}{#1}{#2}\sethlcolor{hlcolor}%
  \hl{#3}%
  \endgroup
}

\title{QQSUM: A Novel Task and Model of Quantitative Query-Focused Summarization for Review-based Product Question Answering}

\author{An Quang Tang \and Xiuzhen Zhang~\thanks{Corresponding author.} \and Minh Ngoc Dinh \and Zhuang Li \\
        RMIT University, Australia \\
        \texttt{s3695273@rmit.edu.vn}, \texttt{xiuzhen.zhang@rmit.edu.au}\\
        \texttt{minh.dinh4@rmit.edu.vn}, \texttt{zhuang.li@rmit.edu.au}
        }

\begin{document}
\maketitle
\begin{abstract}
Review-based Product Question Answering (PQA) allows e-commerce platforms to automatically address customer queries by leveraging insights from user reviews. 
However, existing PQA systems generate answers with only a single perspective, failing to capture the diversity of customer opinions.
In this paper we introduce a  novel task \textbf{Quantitative Query-Focused Summarization (\QQFS)}, which aims to summarize diverse customer opinions into representative Key Points (KPs) and quantify their prevalence to effectively answer  user queries.
While Retrieval-Augmented Generation (RAG) shows promise for PQA, its generated answers still fall short of capturing the full diversity of viewpoints.
To tackle this challenge, our model \textbf{\QQSUMM}, which extends RAG, employs few-shot learning to jointly train a KP-oriented retriever and a KP summary generator, enabling KP-based summaries that capture diverse and representative opinions.  
Experimental results demonstrate that \QQSUMM achieves superior performance compared to state-of-the-art RAG baselines in both textual quality and quantification accuracy of opinions.
Our source code is available at: \url{https://github.com/antangrocket1312/QQSUMM}

\end{abstract}

\section{Introduction}
\label{sec:introduction}
\begin{figure}[tbh]
  \centering
  \includegraphics[width=0.48\textwidth]{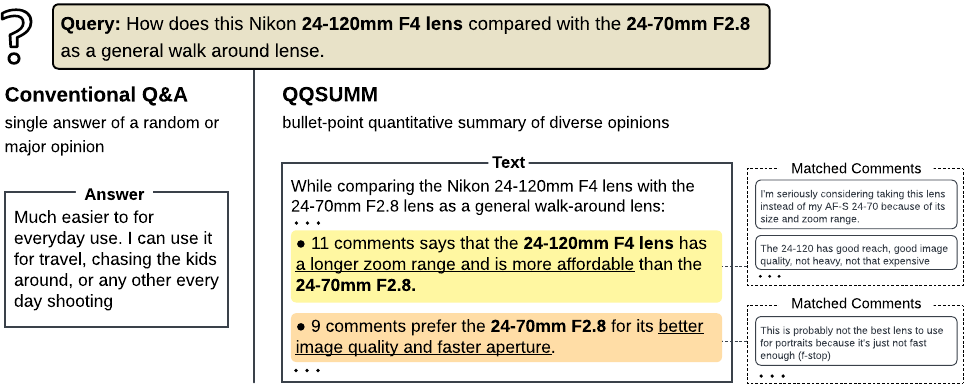}
        \vspace{-2mm}
  \caption{Comparison of conventional Q\&A and \QQFS. More details of \QQFS output are in Table~\ref{tab:qqsumm_example}.}
      \vspace{-3mm}
  \label{fig:QFQS_Overview}
\end{figure}

With the rapid expansion of e-commerce, consumers increasingly rely on product reviews to inform their purchasing decisions. 
Automatic review-based product question answering (PQA) systems have emerged, leveraging user reviews to provide immediate responses on e-commerce Q\&A platforms~\citep{mcauley2016addressing,gupta2019amazonqa}.  
However, current PQA systems face a key limitation: they typically generate a single answer~\citep{gupta2019amazonqa}, overlooking the fact that many subjective e-commerce queries require answers that reflect diverse viewpoints. For example, when comparing camera lenses (Figure~\ref{fig:QFQS_Overview}), some shoppers prioritize versatility and affordability, while others focus on image quality and speed. Recent PQA approaches aim to improve answer quality using retrieval-augmented generation (RAG). These systems first retrieve reviews relevant to the query and then use them as context for large language models (LLMs) to generate answers. Yet, LLMs often struggle to present multifaceted perspectives~\citep{sorensen2024roadmap}, leading to answers that primarily reflect dominant opinions from the retrieved reviews~\citep{deng2020opinion,deng-etal-2023-product}.

Separately, opinion summarization has made progress through Key Point Analysis (KPA), which summarizes reviews into concise, representative statements called key points (KPs) while also quantifying their prevalence~\citep{bar-haim-etal-2020-arguments,bar-haim-etal-2020-quantitative,bar-haim-etal-2021-every,2024aspectbased,tang-etal-2024-prompted}. However, these KPA methods focus on general summarization rather than answering specific queries. For tasks like product comparison, summarization must incorporate only query-focused KPs, making general KPA approaches insufficient for PQA.

In this paper, we introduce a novel task Quantitative Query-Focused Summarization (\QQFS), which generates comprehensive answers containing diverse KPs along with their quantified relative importance (Figure~\ref{fig:QFQS_Overview}). Our solution, \QQSUMM, extends the RAG framework by integrating KP-oriented retrieval and summarization. 
Specifically, \QQSUMM retrieves query-relevant reviews, clusters them by distinct opinions, and summarizes representative KPs from each cluster. This approach provides broader coverage of key insights, overcoming the single-perspective limitation of conventional RAG-based systems.

A key challenge in implementing this approach is  scarcity of training data for such a specialized task. To address this, we develop a co-training strategy that jointly optimizes the retriever and LLM through shared supervision signals, enhancing the alignment between retrieved opinion clusters and generated KPs. This strategy enables robust performance of \QQSUMM even with limited training examples. To support few-shot learning, 
we carefully curated a dataset of queries with KPs and their prevalence quantification, through human-LLM collaboration. Empirical results show that \QQSUMM significantly outperforms RAG baselines based on in-context learning and quantitative summarization.

Our main contributions are:
\begin{itemize}
   \item 
   We introduce a novel task \QQFS.
   Unlike traditional PQA, \QQFS generates answers that capture diverse customer opinions with their prevalence, addressing queries that require multiple viewpoints.
   \item We propose \QQSUMM, a RAG-based framework with KP-oriented retrieval and summarization. The framework is optimized through a co-training strategy that improves alignment between retrieved opinion clusters and generated KPs in few-shot learning setting. Our experiments show that \QQSUMM significantly outperforms baselines with up to 2.11 times improvement in textual similarity with ground-truth KPs and up to 67.12\% improvement in quantification performance over state-of-the-art KPA system for reviews~\citep{tang-etal-2024-prompted}.

\end{itemize}

\section{Related Work}
\label{sec:related_work}
\subsection{Review-based PQA}
Unlike domain-specific QA tasks such as biomedical or legal QA focusing on factual answers,
review-based PQA seeks to provide answers of consumers' subjective opinions about a product.
While extractive PQA approaches retrieve relevant review snippets as answers~\citep{chen2019answer,yu-etal-2012-answering}, it fails to provide precise responses since the review might not be specifically written for answering the given question.
Recently, inspired by the advances of seq-2-seq models, abstractive, i.e., generation-based, approaches can generate natural-language answers from reviews~\citep{chen2019driven,gao2019product}.
However, these approaches frequently suffer from hallucinations and factual inconsistencies, sometimes generating random answers that misrepresent or contradict the prevalent opinions~\citep{deng2020opinion,deng-etal-2023-product}.
Existing review-based PQA framework then cannot capture nor quantify faithfully the diverse opinions of reviews in its answer.

\subsection{Key Point Analysis}
Developed initially to summarize arguments~\citep{bar-haim-etal-2020-arguments,bar-haim-etal-2020-quantitative}, KPA was later adapted for summarization of reviews~\citep{bar-haim-etal-2021-every,2024aspectbased, tang-etal-2024-prompted}.
While ~\citet{bar-haim-etal-2021-every} integrates sentiment analysis and collective key point mining to select and match KPs from broader domain with comments,
~\citet{2024aspectbased} integrates
aspect-based sentiment analysis (ABSA) 
into extracting and matching of KPs to comments 
for more unique KPs and precise quantification.
More recent abstractive  
KPA studies apply abstractive summarization to 
paraphrase and generate KPs from comments (sentences)~\citep{kapadnis-etal-2021-team,li-etal-2023-hear, tang-etal-2024-prompted}.
Overall, whether extractive or abstractive approaches, KPA can only produce KPs for general and high-level opinions
without catering to specific queries.

\subsection{Textual Summarization}
Document summarization aims to produce concise textual summaries capturing the salient information in source documents. 
While extractive review summarization approaches use surface features to rank and extract salient opinions for summarization~\citep{mihalcea-tarau-2004-textrank,angelidis-lapata-2018-summarizing,zhao2020weakly}, 
abstractive techniques use sequence-to-sequence models~\citep{chu2019meansum,suhara-etal-2020-opiniondigest,brazinskas-etal-2020-unsupervised,brazinskas-etal-2020-shot,zhang2020pegasus} to generate review-like summaries containing only the most prevalent opinions.
Recently, prompted opinion summarization leveraging Large Language Models (LLMs) was applied to generate fluent and concise review summaries~\citep{bhaskar2023prompted}.
However, existing studies lack focus on presenting and quantifying the diverse opinions in reviews.

\section{Quantitative Query-Focused Summarization}
\label{sec:methodology}
\subsection{Task Formulation}
\label{sec:task}

Let $q$ denote a query, i.e., community question, and \mbox{$R_e = \{r_j\}_{j=1}^{|R_e|}$} denotes a set of review comments on a product $e$, \QQFS aims to retrieve relevant comments $\mathcal{D}$ to answer $q$ and generate a KP-based summary $\mathcal{S}$ quantifying viewpoints presented in $\mathcal{D}$.
We formulate $\mathcal{S}=\{kp_1,\ldots,kp_n\}$ as a bullet-point summary containing multiple KPs, where each bullet-point represents a KP~
\footnote{unique and non-overlapping opinion at high level} and its prevalence~\citep{bar-haim-etal-2021-every}. 
For instance, with the bullet-point ``\emph{23 comments praise that the headphone is very comfortable for long hours}'', the KP is ``\emph{Comfortable for long hours}'', and the prevalence count is 23.
Each key point $kp_i$, is matched to a subset of supporting comments $\mathcal{C}_i=\{c_1,c_2,\ldots\}$ (where $c_i \in \mathcal{D}$), with prevalence being measured as $|\mathcal{C}_i|$.

\subsection{The \QQSUMM Framework}
\begin{figure*}[tbh]
  \centering
  \includegraphics[width=0.85\textwidth]{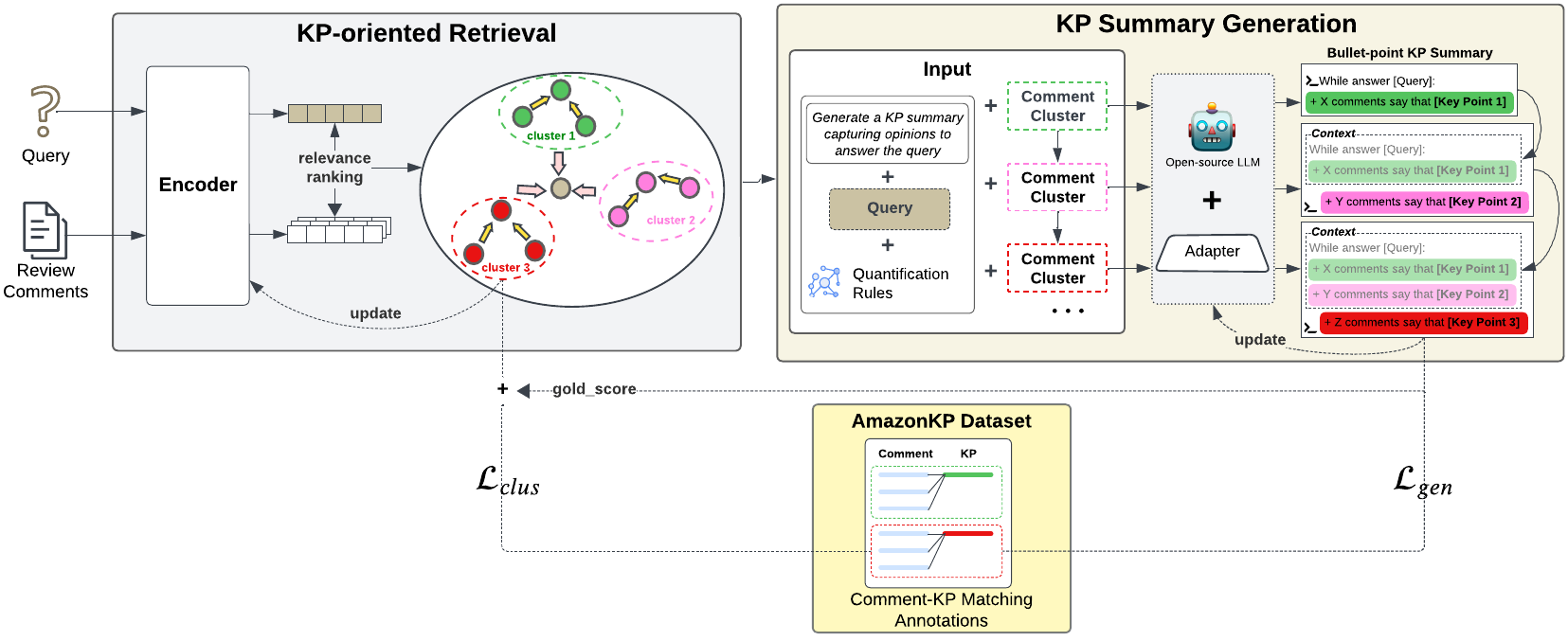}
          \vspace{-2mm}
  \caption{The training architecture of the \QQSUMM framework.
  }
  \vspace{-4mm}
  \label{fig:QQSumm_Framework}
\end{figure*}

Figure~\ref{fig:QQSumm_Framework} illustrates the architecture of \QQSUMM.
\QQSUMM is based on the retrieval-augmented generation (RAG) paradigm and consists of 2 stages: 
\textbf{\emph{KP-Oriented Retrieval}} and \textbf{\emph{KP Summary Generation}}.
It utilizes a Retriever to retrieve and cluster query-relevant comments into groups, and the LLM to generate the final KP summary based on the comment clusters.
Importantly, the retriever and LLM can be jointly trained with shared supervision signals to ensure comment clusters retrieved match KPs generated.

The following general loss function describes every training step of \QQSUMM, whose parameters are updated
at the cluster-KP level rather than the query level:
\begin{equation}\label{equation-key-point-generation}
\small
\mathcal{L} = (1-d) \cdot (\mathcal{L}_{\text{clus}} + \text{gold\_score}) + d \cdot \mathcal{L}_{\text{gen}}
\end{equation}
where $\mathcal{L}_{clus}$ is the retrieval loss for each comment cluster, $\mathcal{L}_{gen}$ is the LLM’s generation loss computed for the KP generated from the respective cluster, and $d$ is a damping factor to balance between the two.
Notably, $gold\_score$ represents the Perplexity Distillation loss~\citep{izacard2023atlas}, which transforms the supervisory signals of the LLM to improve the Retriever.
The intuition is that within a cluster, comments that better contribute to helping the LLM generate the KP with lower perplexity should be ranked higher.

\subsubsection{KP-Oriented Retrieval}
Given a query $\mathbf{q}$, the Retriever should retrieve relevant review comments $R_q$ that emphasize opinions focused on $\mathbf{q}$.
We utilize a shared encoder $\textbf{E}$ that can encode both the input query $\mathbf{q}$ and each review comment $\mathbf{r}_{j} \in R_e$. 
Comments are ranked by the similarity score 
$s(\mathbf{x},\mathbf{r}_{j})=\textbf{E}_c(\mathbf{x})^\top\textbf{E}_d(\mathbf{r}_{j})$ 
that is calculated by taking the dot product of the embeddings of the query $\mathbf{x}$ and the comment $\mathbf{r}_{j}$.
Only comments with $s(\mathbf{x},\mathbf{r}_{j}) \geq 1$ is selected for $R_q$.~\footnote{the similarity threshold 1 is set empirically} 

Different from standard RAG where generation is based on the direct retrieval result, to ensure diverse and representative opinions for generation, we enhance the Retriever with the clustering objective to produce distinctive comment groups that conceptually match KPs for generation.

\paragraph{KP-Oriented Retrieval Loss}
Starting with an empty list of clusters $\mathbf{C}$, and iterate through every comment in $R_q$, for every comment, we further iterate through every existing cluster $\mathbf{c}_i \in \mathbf{C}$ and calculate its average cosine similarity score to all comments of the cluster.
Finally, we add the comment to any clusters with average cosine similarity score above a threshold ($\lambda = 1.2$),~\footnote{set empirically based on cluster quality} otherwise, a new cluster is created.
Importantly, a comment can be mapped to multiple clusters.
We empirically showed that our proposed clustering algorithm is more effective than HDBSCAN~\citep{mcinnes2017hdbscan} and K-Means through an ablation experiment in Appendix~\ref{sec:ablation_clustering_algorithm_eval}.

To train the retriever for KP-oriented retrieval, we align predicted comment clusters $\mathbf{C}$ with annotated clusters $\mathbf{P}$, where $\mathbf{P}$ groups comments matched to the same KP (annotation details in §\ref{sec:amazonkp}). 
The centroid embedding of a cluster is the mean embedding of its comments $\mathbf{r}_k$: 
$\bar{\textbf{E}}_{c}(\mathbf{c}_i)=\frac{1}{M}\sum_{k=1}^{M} {\textbf{E}}(\mathbf{r}_{k})$.
Because a cluster $\mathbf{c}_i \in \mathbf{C}$ 
may contain mixed opinions represented by 
multiple clusters from $\mathbf{P}$, we map each $\mathbf{c}_i$ to the mean embedding of 
$\mathbf{P}_{match} \subset \mathbf{P}$: 
$\bar{\textbf{E}}_{c}(\mathbf{P}_{match})=\frac{1}{M}\sum_{j=1}^{M} \bar{\textbf{E}}_{c}(\mathbf{p}_{j})$, 
where the semantic similarity between $\mathbf{c}_i$ and every $\mathbf{p}_j$ is 
$sim(\mathbf{c}_i,\mathbf{p}_{j})=\bar{\textbf{E}}_{c}(\mathbf{c}_i)^\top\bar{\textbf{E}}_{c}(\mathbf{p}_j) \geq threshold$.
The training objective minimizes the mean-squared-error (MSE) loss between each comment $\mathbf{r}_{k}$ in $\mathbf{c}_i$ and the average center of the most similar clusters $\mathbf{P}_{match}$.
\begin{equation}  
  \mathcal{L}_{clus}=
  \frac{1}{|\mathbf{c}_i|}\sum_{k=1}^{|\mathbf{c}_i|} || \bar{\textbf{E}}_{c}(\mathbf{P}_{match})-\textbf{E}(\mathbf{r}_{k})||_2^2.
\end{equation}

\subsubsection{KP Summary Generation}

A key limitation of previous KPA studies is that KPs may contain redundant opinions, due to that review comments,
possibly containing multiple opinions, 
are mapped to individual KPs locally~\citep{bar-haim-etal-2021-every,tang-etal-2024-prompted}.
To address this limitation, we propose to generate KPs at the global level, where the goal is to generate an overall KP-based summary without redundancy. 
Our main idea is that generated KPs are used as the context for the LLM to better reason and generate the next KP, which should be a unique, non-overlapping opinion statement.

\paragraph{Prompting Strategies}
Following OpenAI's prompt engineering guidelines\footnote{\url{https://platform.openai.com/docs/guides/prompt-engineering}}, we format query-relevant comment clusters from the Retriever into a structured prompt with four parts (detailed in Listing~\ref{lst:few_shot_kpsg}, Appendix~\ref{sec:kpsg_prompt}): 
\textbf{1)} Context and input structure, 
\textbf{2)} Task definition and output requirements, 
\textbf{3)} Summarization steps for identifying representative KPs per cluster and generating the final KP-based summary, and 
\textbf{4)} Commonsense quantification rules 
to prioritize clusters by size and prevent overlapping KPs. 
To minimize ambiguity and hallucination, we encode predicted clusters $\mathbf{C}$ as JSON objects and assign each a unique ID, requiring the LLM to label generated KPs accordingly.

\paragraph{Next-KP-Generation Training}
During training, generating multiple KPs in a summary lacks alignment with $\mathcal{L}_{clus}$, which is computed per comment cluster.
To address this, we introduce a \emph{Next-KP-Generation} objective, inspired by Next-Token Prediction in LMs~\citep{NEURIPS2020_1457c0d6}, to enhance the generation of salient, non-overlapping KPs. This approach fine-tunes the LLM to iteratively generate KPs within the summary.
Specifcally, let the final KP-based summary $\mathcal{S}=\{kp_1,\ldots, kp_i, \ldots, kp_n\}$, 
each $kp_i$ is generated with preceding KPs $\{kp_1,\ldots, kp_{i-1}\}$ as the context, prompting the LLM to iteratively complete $\mathcal{S}$. 
The generation loss for each $kp_i$ of $\mathbf{c}_i \in \mathbf{C}$ is computed as the negative log-likelihood (NLL) against the reference KP, annotated for the most similar $\mathbf{p}_i \in \mathbf{P}$ identified during retrieval,

\begin{equation}
  \mathcal{L}_{gen} = -\frac{1}{T} \sum_{t=1}^{T} \log P(x_t | x_{<t})
\end{equation}
where \( P(x_t | x_{<t}) \) represents the probability assigned by the model to the correct token \( x_t \), given the preceding tokens \( x_{<t} \).

\subsection{Human-LLM Key Point Annotation}
\label{sec:amazonkp}
From Section~3.2, to train our \QQSUMM framework in the few-shot setting, 
annotation of KPs for queries and relevant comments are necesary. 
Prior KPA studies only include annotations matching comments to KPs without queries~\citep{bar-haim-etal-2020-arguments,bar-haim-etal-2020-quantitative}.
No datasets exist for matching comments to KPs in PQA.

\begin{figure}[tbh]
  \centering
  \includegraphics[width=0.48\textwidth]{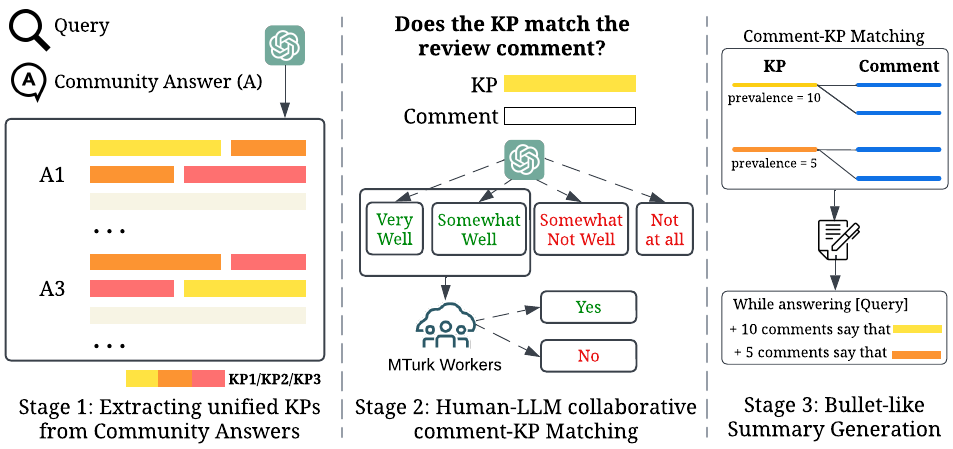}
          \vspace{-2mm}
  \caption{Illustration of the human-LLM collaborative annotation pipeline for \AmazonKP.
  }
          \vspace{-4mm}
  \label{fig:QQSumm_Annotation}
\end{figure}

\begin{table}[]
  \centering
  \small
  \scalebox{0.9}{
  \begin{tabularx}{0.5\textwidth}{@{}l|X|X@{}}
  \toprule
  \textbf{Statistic} & \multicolumn{1}{l}{Train} & \multicolumn{1}{l}{Test} \\ \midrule
  \# Product Categories & 17 & 17 \\
  \# Instances (queries) Per Category & 2 & 148 \\
  Total Instances & 34 & 2516 \\
  \# Reviews Per Query & 71.18 & 72.70 \\
  \# Review Comments Per Query & 452.03 & 431.62 \\
  \# Answers Per Query & 7.53 & 6.45 \\ \hdashline
  \# KPs Per Query (Stage 1) & 9.26 & 6.90 \\
  \# Relevant Comments Per Query (Stage 2) & 24.50 & -- \\
  \# Comments (Prevalence) per KP (Stage 2) & 6.37 & -- \\
  Summary Length (Stage 3) & 101.29 & -- \\ \bottomrule
  \end{tabularx}
  }
    \vspace{-2mm}
  \caption{Core statistics of the \AmazonKP dataset. 
  }
  \vspace{-5mm}
  \label{tab:AmazonKP_stats}
\end{table}

We leverage the popular PQA dataset AmazonQ\&A~\citep{gupta2019amazonqa} for our \QQFS task, 
focusing on only \emph{answerable}, \emph{subjective} (non-factual) questions that have multiple answers.
Out of 17 product categories (e.g., Electronics, Video Games),  
we only include businesses with 50-100 reviews,
and sampling top 150 questions per category
based on answer count. 
For ease of reference we name this curated dataset \AmazonKP.
Details on question classification for \AmazonKP are in Appendix~\ref{sec:opinionated_question_classification},
and their taxonomy in Appendix~\ref{sec:qual_analysis_question_amazonkp}. 
Notably, the dominance of ``\emph{Scenario-based}'' questions underscore the importance of \QQFS for generating KP summary to answer user questions on preferences and scenarios.

Manually summarizing and quantifying opinions from comments is laborious and time-consuming, if not impossible.
Research shows LLM's strong annotation capabilities~\citep{he-etal-2024-annollm}, and so we design a three-stage human-LLM collaborative annotation pipeline, shown in Figure~\ref{fig:QQSumm_Annotation}.

\paragraph{Stage 1: KP Extraction from Gold Community Answers}
Given a query $q_i$, the AmazonQ\&A dataset provides multiple answers, i.e. responses, from online users 
$A_i=\{a_1, a_2, \ldotp\}$, serving as ideal approximation of gold opinions. 
However, these responses can contain overlapping opinions.
We therefore zero-shot prompted \texttt{GPT-4-o-mini} to extract distinctive and non-overlapping KPs from $A_i$.
Empirical validation with human annotators confirms that the extracted KPs are of high quality, 
with 90\% of community answers were represented by KPs, 
while 87.5\% of the extracted KPs are verified as valid (precision).
Further details are in Appendices~\ref{sec:kp_extraction_answer_validation} and~\ref{sec:kp_extraction_answer}.

\paragraph{Stage 2: LLM-based and Manual Comment-KP Matching}
Based on the annotation process in the literature~\citep{bar-haim-etal-2020-arguments}, we further integrate LLMs to reduce human effort and time. Using KPs extracted from gold answers (Stage~1), we prompt \texttt{GPT4-o-mini} to annotate pairwise matches between comments and KPs from all available reviews of the product. LLM-matched pairs are then validated by three Amazon Mechanical Turk (MTurk) workers. Finally, comments from validated pairs are grouped by similar KPs, with KP prevalence determined by the number of matching comments.
Further details on KP Matching annotations are provided in Apppendix~\ref{sec:annotation_detail_kp_matching}.

\paragraph{Stage 3: KP-based Summary}
We utilize KPs and their prevalence counts, discovered for every query, to manually compose a bullet-point KP-based summary, where each bullet point corresponds to a KP and is annotated as ``$|kp_i|$ \emph{comments say that} $kp_i$''.

The number of pairwise comment-KP matching annotations required per query can be up to 2K-3.5K. \textbf{For training}, to control annotation costs, we conducted Stages 1, 2 and 3 annotations on a small subset of 34 instances for few-shot training of \QQSUMM, randomly selecting two queries per product category for supervised labeling. \textbf{For evaluating the KP-based summary}, the remaining examples with only Stage 1 annotations serve as the test set. 
The core statistics of \AmazonKP are shown in Table~\ref{tab:AmazonKP_stats}.

\section{Experiments}
\label{sec:experiment}
We employ Atlas~\citep{izacard2023atlas}, a pre-trained efficient RAG model, as our backbone model for \QQSUMM. We utilized Contriever~\citep{izacard2022unsupervised} as the retriever while replacing the original language model with open-source LLMs 
(e.g., \texttt{Vicuna-7B}~\footnote{\url{https://huggingface.co/lmsys/vicuna-7b-v1.5}}, 
\texttt{Mistral-7B}~\footnote{\url{https://huggingface.co/mistralai/Mistral-7B-Instruct-v0.2}}) for generation. 
For computational feasiblity, we apply Low-Rank Adaptation (LoRA)~\citep{hu2021lora}, which adds trainable parameters while freezing the model’s original weights.

\subsection{Baselines}
We benchmark \QQSUMM against 3 RAG baselines.
\paragraph{(Retriever + LLM)\textsubscript{co-train}}
We few-shot trained Atlas~\citep{izacard2023atlas}, with the standard RAG architecture and Retriever-LLM generator co-training, for the \QQFS task.
The retriever retrieves relevant comments, while letting the LLM implicitly infer KPs' matching comments and their quantities during KP summary generation.
For training, we aggegrated matching comments across KPs, per query, as the retrieval ground truth.

\paragraph{Frozen Retriever + Prompt LLM}
To assess in-context learning (ICL) for \QQFS, we use a frozen retriever and \texttt{Vicuna-7B}, \texttt{Mistral-7B}, and \texttt{GPT-4-Turbo} as the LLM for ICL. Few-shot training instances are concatenated with test 
instances, with the number of few-shot examples optimized for context length and cost: 4-shot for \texttt{Mistral-7B} and \texttt{GPT-4-Turbo}, and 2-shot for \texttt{Vicuna-7B}.

\paragraph{Frozen Retriever + KPA}
We replace the LLM of a standard RAG with existing KPA review summarization systems to adapt KPA to the \QQFS task.
Comments were first retrieved by a frozen retriever and then \textbf{RKPA-Base}~\citep{bar-haim-etal-2021-every} utilizes a quality ranking model~\citep{gretz2020large} to extract KP candidates before matching comments to KPs using a KP Matching model~\citep{bar-haim-etal-2020-quantitative} at threshold $t_{match} = 0.99$.
\textbf{PAKPA}~\citep{tang-etal-2024-prompted} clusters comments by aspect and sentiment before generating aspect-oriented KPs.

All experiments were conducted at the KP level, focusing on KPs in the summary outputs of \QQSUMM and baselines for fair comparison. 
We post-process the output KP-based summary into 
KPs as JSON objects, where each object covers the KP information of a bullet point in the summary.~\footnote{We use a simple LLM-based post processor, prompting \texttt{gpt-4-o-mini} with 
\texttt{'Format all key points and their prevalences mentioned in the above bullet-point summary in a JSON list, where each JSON object format as: \{'key\_point': <key point of a bullet>, 'prevalence': <key point's prevalence>\}'}}
The baselines were implemented using either the PyTorch module or the Huggingface transformers framework, and were trained on a NVIDIA GeForce RTX 4090 GPU.

\subsection{Evaluation Dimensions}
\label{sec:evluation_dimensions}
We conducted experiments on the test set of \AmazonKP (§\ref{sec:amazonkp}), consisting of questions from 17 product categories. For reasonable cost, we sample 8 questions from each category for evaluation.

\subsubsection{KP Textual Quality}
\paragraph{Automatic Evaluation}
Extracted KPs from gold community answers in AmazonKP (Stage 1 of §~\ref{sec:amazonkp}) serves as the reference KPs for this automatic evaluation.
We first perform a \emph{lexical} comparison between KPs in the generated summary and the ground truth by computing the highest $\mathtt{Rouge}$~\citep{lin-2004-rouge} score between generated and reference KPs for each query and then average the maxima.
Then, following ~\citet{li-etal-2023-hear}, we calculate soft-Precision/Recall/F1~(denoted by sP, sR and sF1, respectively), 
which measure the \emph{semantic} similarity between individual generated KP and the reference KP.
While $sP$ finds the reference KP with the highest similarity score for each generated KP, $sR$ is vice-versa, and ($sF1$) is the harmonic mean between $sP$ and $sR$.
\begin{equation}
  \small
  sP = \frac{1}{n} \times \sum_{ \alpha_i\in\mathcal{A}} \max_{\beta_j\in\mathcal{B}} f(\alpha_{i},  \beta_{j})
\end{equation}
\begin{equation}
  \small
  sR = \frac{1}{m} \times \sum_{ \beta_i\in\mathcal{B}} \max_{\alpha_j\in\mathcal{A}} f(\alpha_{i},  \beta_{j})
\end{equation}

Additionally, inspired by sP/sR/sF1 of ~\citet{li-etal-2023-hear}, we further propose $RD$ to identify KP \emph{redundancy}. 
For each generated KP in the summary for a query, $RD$ finds the neighborhood KP with the highest similarity score.
\begin{equation}
  \small
  RD = \frac{1}{n} \times \sum_{ \alpha_i\in\mathcal{A}} \max_{\theta_j\neq\alpha_i\in\mathcal{A}} f(\alpha_{i},  \theta_{j})
\end{equation}

where $f$ computes similarities between two individual key points, $\mathcal A$, $\mathcal{B}$ is the set of generated and reference KPs and $n=|\mathcal{A}|$ and $m=|\mathcal{B}|$, respectively.
We use state-of-the-art semantic similarity metrics $\mathtt{BLEURT}$~\citep{sellam-etal-2020-bleurt} and $\mathtt{BERTScore}$~\citep{Zhang*2020BERTScore}, along with LLM-based metric \textsc{G-Eval}-4~\citep{liu-etal-2023-g}
as $f_{max}$.
Note that \textsc{G-Eval} scores are scaled from 1-5 to 0-1 for comparability and its evaluation prompt was also customized to fit our evaluation (Appendix~\ref{sec:g_eval_prompt}).

\paragraph{Human Evaluation}
We manually evaluated the information quality of generated KPs in the summary considering 7 different dimensions utilized in previous KPA studies~\citep{kapadnis-etal-2021-team,tang-etal-2024-prompted}, 
including \textsc{Redundancy}, \textsc{Coverage}, \textsc{Faithfulness} \textsc{Validity}, \textsc{Sentiment}, \textsc{Informativeness} and \textsc{Single\ Aspect}.
Details of these dimensions are in Appendix~\ref{sec:kp_quality_dimensions}.

We conducted pairwise comparisons of KPs from different systems using Amazon Mechanical Turk (MTurk). Given a dimension for evaluation, each comparison involved choosing the better one from two summaries, each taken from a different system.
Using the Bradley-Terry model~\citet{friedman-etal-2021-overview}, we calculated rankings from these comparisons among the models. 
For an example of an annotation, see Appendix~\ref{sec:pairwise_kp_quality_annotation_guideline}.
Note that for reasonable cost, we sample and select only the popular question (with the highest average KP prevalence), 
each from 5 common categories~\footnote{namely \emph{Home \& Kitchen}, \emph{Sports \& Outdoors},
\emph{Tools \& Home Improvement}, \emph{Health \& Personal Care}, and \emph{Beauty}} of \textsc{AmazonKP}.

\subsubsection{KP Quantification Performance}
\label{sec:kp_quantification}
We evaluate the KP quantification performance of different systems for KP-comment matching and factual alignment.

\paragraph{KP-comment matching}
We first assess the accuracy of the KP comment matching by extending \citet{bar-haim-etal-2021-every} to measure both \emph{precision} (correctness of predicted matches) and \emph{recall} (coverage of ground-truth matches). For each system, we compute precision and recall by prompting \texttt{gpt-4-o-mini} to annotate pairwise \emph{match}/\emph{non-match} between generated KPs and retrieved comments $R_q$. Additionally, leveraging annotated comment-KP pairs, we introduce \emph{QuantErr}, which measures the mean absolute error between predicted and actual KP prevalence count. 
Empirical validation shows \texttt{gpt-4-o-mini} annotations highly correlated with MTurk workers' judgement (Pearson’s $r$ = 0.647) (Appendix~\ref{sec:gpt4_annotation_validate}).

\paragraph{KP-comment factual alignment:}
We further employed $\mathtt{AlignScore}$~\citep{zha-etal-2023-alignscore} for automatic evaluation of factual alignment between generated KPs and their corresponding comments.

\begin{table}[t]
  \centering
  \fontsize{8pt}{8pt}\selectfont
  \begin{tabularx}{0.5\textwidth}{Xcccc}
  \toprule
  & P@5 & P@10 & P@20 & P@all \\
  \midrule
  \rowcolor{lightgray}
  \multicolumn{5}{l}{$\mathbf{\QQSUMM}$ (Ours)}\\
  \specialrule{0em}{1pt}{1pt}
  \multicolumn{5}{l}{Contriever~\citep{izacard2022unsupervised}}\\
  \hspace{3mm} + Mistral & \textbf{0.668} & \textbf{0.633} & \textbf{0.601} & \textbf{0.535} \\
  \hspace{3mm} + Vicuna & 0.567 & 0.527 & 0.526 & 0.367 \\
  \multicolumn{5}{l}{all-MiniLM-L12-v2~\citep{wang-etal-2021-minilmv2}}\\
  \hspace{3mm} + Mistral & 0.590 & 0.538 & 0.500 & 0.440 \\
  \hspace{3mm} + Vicuna & 0.569 & 0.507 & 0.468 & 0.362 \\
  \rowcolor{lightgray}
  \rowcolor{lightgray}
  \multicolumn{5}{l}{$\mathbf{(Retriever + LLM)\textsubscript{co-train}}$~\citep{izacard2023atlas}}\\
  \specialrule{0em}{1pt}{1pt}
  \multicolumn{5}{l}{Contriever~\citep{izacard2022unsupervised}}\\
  \hspace{3mm} + Mistral &  0.544 & 0.511 & 0.459 & 0.345 \\
  \hspace{3mm} + Vicuna & 0.444 & 0.467 & 0.442 & 0.328 \\
  \multicolumn{5}{l}{all-MiniLM-L12-v2~\citep{wang-etal-2021-minilmv2}}\\
  \hspace{3mm} + Mistral & 0.531 & 0.530 & 0.515 & 0.350 \\
  \hspace{3mm} + Vicuna & 0.552 & 0.512 & 0.454 & 0.339 \\
  \rowcolor{lightgray}
  \multicolumn{5}{l}{$\mathbf{frozen\ Retriever + prompt\ LLM}$}\\
  \specialrule{0em}{1pt}{1pt}
  Contriever & 0.494 & 0.447 & 0.404 & 0.325 \\
  all-MiniLM-L12-v2 & 0.479 & 0.446 & 0.452 & 0.315 \\
  BM25 & 0.469 & 0.432 & 0.387 & 0.283 \\
  \bottomrule
  \end{tabularx}
    \vspace{-2mm}
  \caption{\label{table:retrieval_performance}
  Performance of retrieval models.}
      \vspace{-5mm}
\end{table}

\subsection{Results}
\begin{table*}[t]
  \centering
  \scalebox{0.65}{
  \begin{tabular}{lccccccccccccccc}
  \toprule
  {} & \multicolumn{3}{c}{ROUGE} & \multicolumn{4}{c}{BERTScore} & \multicolumn{4}{c}{BLEURT} & \multicolumn{4}{c}{G-Eval-4} \\
  \cmidrule(r){2-4} \cmidrule(r){5-8} \cmidrule(r){9-12} \cmidrule(r){13-16}
   & R-1 & R-2 & R-L & sP & sR & sF1 & RD$\downarrow$ & sP & sR & sF1 & RD$\downarrow$ & sP & sR & sF1 & RD$\downarrow$ \\
  \midrule
  \rowcolor{lightgray}
  \multicolumn{16}{l}{$\mathbf{QQSUM{-}RAG}$ (Ours)}\\
  \specialrule{0em}{1pt}{1pt}
  \hspace{3mm} + Mistral & \textbf{0.256} & 0.061 & \textbf{0.220} & \textbf{0.39} & \textbf{0.29} & \textbf{0.33} & \textbf{0.37} & \textbf{0.51} & \textbf{0.41} & \textbf{0.46} & \textbf{0.49} & \textbf{0.88} & \textbf{0.82} & \textbf{0.85} & \textbf{0.36} \\
  \hspace{3mm} + Vicuna & 0.222 & \textbf{0.078} & 0.204 & 0.38 & 0.26 & 0.31 & 0.53 & 0.49 & 0.39 & 0.44 & 0.54 & 0.87 & 0.81 & 0.84 & 0.36 \\ 
  \rowcolor{lightgray}
  \multicolumn{16}{l}{$\mathbf{(Retriever + LLM)_{co{-}train}}$~\citep{izacard2023atlas}}\\
  \specialrule{0em}{1pt}{1pt}
  \hspace{3mm} + Mistral & 0.209 & 0.057 & 0.194 & 0.37 & 0.28 & 0.32 & 0.43 & 0.49 & 0.40 & 0.44 & 0.55 & 0.81 & 0.82 & 0.81 & 0.41 \\
  \hspace{3mm} + Vicuna & 0.174 & 0.041 & 0.161 & 0.37 & 0.26 & 0.31 & 0.48 & 0.48 & 0.38 & 0.42 & 0.58 & 0.78 & 0.78 & 0.78 & 0.41 \\
  \rowcolor{lightgray}
  \multicolumn{16}{l}{$\mathbf{Frozen\ Retriever + prompt\ LLM}$}\\
  \specialrule{0em}{1pt}{1pt}
  \hspace{3mm} + Mistral & 0.210 & 0.055 & 0.191 & 0.33 & 0.26 & 0.29 & 0.51 & 0.46 & 0.38 & 0.42 & 0.55 & 0.79 & 0.80 & 0.79 & 0.41 \\
  \hspace{3mm} + Vicuna & 0.164 & 0.059 & 0.154 & 0.22 & 0.20 & 0.21 & 0.48 & 0.46 & 0.31 & 0.37 & 0.59 & 0.73 & 0.73 & 0.73 & 0.41 \\
  \hspace{3mm} + GPT-4-Turbo & 0.197 & 0.048 & 0.174 & 0.32 & 0.25 & 0.28 & 0.44 & 0.45 & 0.38 & 0.41 & 0.54 & 0.77 & 0.77 & 0.77 & 0,38 \\
  \rowcolor{lightgray}
  \multicolumn{16}{l}{$\mathbf{Frozen\ Retriever + KPA}$}\\
  \specialrule{0em}{1pt}{1pt}
  \hspace{3mm} + PAKPA~\citep{tang-etal-2024-prompted} & 0.179 & 0.027 & 0.162 & 0.34 & 0.28 & 0.31 & 0.46 & 0.47 & 0.41 & 0.44 & 0.54 & 0.79 & 0.80 & 0.80 & 0.36 \\
  \hspace{3mm} + RKPA-Base~\citep{bar-haim-etal-2021-every} & 0.121 & 0.016 & 0.106 & 0.16 & 0.14 & 0.14 & 0.50 & 0.43 & 0.36 & 0.39 & 0.61 & 0.69 & 0.70 & 0.69 & 0.51 \\
  \bottomrule
  \end{tabular}
  }
  \caption{\label{table:automatic_evaluation}
  KP summary textual quality.
  sP, sR and sF1 refer to Soft-Precision, Soft-Recall, and Soft-F1 respectively based on set-level evaluation method against 
  reference KPs in gold answer.
  }
  \vspace{-3mm}
\end{table*}

\begin{table*}[htbp]
  \centering
  \scalebox{0.65}
  {
  \begin{tabular}{lccc:cccc}
      \toprule
      & CV & FF & RD & VL & SN & IN & SA\\
      \midrule
      \QQSUMM (Ours) & \textbf{28.44} & \textbf{26.56} & \textbf{25.34} & \textbf{35.23} & \textbf{31.11} & \textbf{25.9} & \textbf{24.8} \\
      (Retriever + LLM)\textsubscript{co-train}~\citep{izacard2023atlas} & 11.06 & 11.17 & 14.7 & 9.99 & 9.54 & 13.49 & 17.52 \\
      Frozen Retriever + prompt LLM (GPT-4-Turbo) & 15.12 & 12.84 & 15.73 & 10.36 & 14.6 & 12.59 & 10.79 \\
      \hdashline
      Frozen Retriever + PAKPA~\citep{tang-etal-2024-prompted} & 9.94 & 12.41 & 13.28 & 7.7 & 8.87 & 13.04 & 9.34\\
      Frozen Retriever + RKPA-Base~\citep{bar-haim-etal-2021-every} & 16.20 & 22.28 & 15.73 & 22.91 & 20.75 & 21.02 & 18.77 \\
      \bottomrule
  \end{tabular}
  }
  \caption{Human evaluation of KP information quality by different dimensions.
  Reported are the Bradley Terry scores of 7 dimensions, from left to right, \textsc{Coverage}, \textsc{Faithfulness} and \textsc{Redundancy}, \textsc{Validity}, \textsc{Sentiment}, \textsc{Informativeness}, \textsc{SingleAspect}. 
  For reasonable cost, we only conducted manual evaluation on \texttt{Mistral} - the best LM configuration of \QQSUMM and (Retriever + LLM)\textsubscript{co-train}, selected from Table~\ref{table:automatic_evaluation}.
  \label{table:kp_quality_eval_results_T}}
\end{table*}

\begin{table*}[t]
  \centering
  \scalebox{0.65}{
  \begin{tabular}{lccccc}
  \toprule
  {} & \multicolumn{4}{c}{KP-Comment Matching} & \multicolumn{1}{c}{\makecell{KP-Comment Factual Alignment}} \\
  \cmidrule(r){2-5} \cmidrule(r){6-6}
  & P & R & F1 & QuantErr$\downarrow$ & AlignScore \\
  \midrule
  \rowcolor{lightgray}
  \multicolumn{6}{l}{$\mathbf{\QQSUMM}$ (Ours)}\\
  \specialrule{0em}{1pt}{1pt}
  \hspace{3mm} + Mistral & 0.694 & \textbf{0.869} & \textbf{0.792} & \textbf{04.24} & \textbf{0.749} \\
  \hspace{3mm} + Vicuna & 0.538 & 0.684 & 0.602 & 07.83 & 0.630 \\
  \rowcolor{lightgray}
  \multicolumn{6}{l}{$\mathbf{(Retriever + LLM)_{co{-}train}}$~\citep{izacard2023atlas}}\\
  \specialrule{0em}{1pt}{1pt}
  \hspace{3mm} + Mistral & 0.567 & 0.249 & 0.346 & 18.10 & 0.653 \\
  \hspace{3mm} + Vicuna & 0.442 & 0.094 & 0.154 & 30.13 & 0.394 \\
  \rowcolor{lightgray}
  \multicolumn{6}{l}{$\mathbf{Frozen\ Retriever + prompt\ LLM}$}\\
  \specialrule{0em}{1pt}{1pt}
  \hspace{3mm} + GPT-4-Turbo & 0.746 & 0.200 & 0.313 & 16.63 & 0.673 \\
  \hspace{3mm} + Mistral & 0.498 & 0.214 & 0.300 & 19.14 & 0.624 \\
  \hspace{3mm} + Vicuna & 0.439 & 0.185 & 0.260 & 21.52 & 0.531 \\
  \rowcolor{lightgray}
  \multicolumn{6}{l}{$\mathbf{Frozen\ Retriever + KPA}$}\\
  \specialrule{0em}{1pt}{1pt}
  \hspace{3mm} + PAKPA~\citep{tang-etal-2024-prompted} & \textbf{0.762} & 0.520 & 0.619 & 06.68 & {\bf 0.749} \\
  \hspace{3mm} + RKPA-Base~\citep{bar-haim-etal-2021-every} & 0.371 & 0.314 & 0.340 & 15.62 & 0.354 \\
  \bottomrule
  \end{tabular}
  }
  \vspace{-3mm}
  \caption{\label{table:quantitative_evaluation}
  Performance for KP-Comment matching and factual alignment
  }
  \vspace{-5mm}
\end{table*}

\subsubsection{The Retrieval Model}
The retriever is important for retrieving comments relevant to queries and so we first evaluated the performance of different backbone retrieval models.
For this we prompted \texttt{gpt-4-o-mini} to annotate the relevance of retrieved comments to queries.
Table~\ref{table:retrieval_performance} reports the retrieval Precision@k (P@k), measured at different levels of top-k-ranked retrieved comments ($[5, 10, 20, all]$), using 3 different retrieval models: 
Contriever, all-MiniLM-L12-v2 and BM25.
Note that BM25 is not a neural encoder and therefore can only be evaluated in the frozen Retriever + Prompt LLM setting.

Overall, the trained retriever of \QQSUMM, as being co-trained with the LLM and extended for KP-oriented retrieval, outperform all baselines.
Notably, co-training with stronger LM can also contribute up to 45.78\% improvement, as the supervision signal from more query-focused KP generation helps train the Retriever to rank documents more accurately.
Contriever stood out as the best performer regardless of the LM selection.
Hereafter we base all upcoming experiments with Contriever as the retrieval model.

\subsubsection{KP Quality}
KPs produced by different systems in terms of textual quality, semantic quality and redundancy are reported in Table~\ref{table:automatic_evaluation}.
Scores of all systems are low in general, as opinions in product reviews may not cover all opinions from community answers to questions.
From Table~\ref{table:automatic_evaluation}, \QQSUMM outperforms other systems in all quality dimensions.
It shows 2.11 times improvement in textual similarity with reference KPs (0.256 vs. 0.121 in ROUGE-1),
0.23 point absolute improvement in semantic similarity (0.39 vs. 0.16 in BERTScore)
and 0.14 point absolute reduction in Redundancy (0.37 vs. 0.51 using BERTScore for semantic similarity).

The high quality of KPs in \QQSUMM can be attributed to  the KP-oriented retrieval of \QQSUMM.
Notably, although (Retriever + LLM)\textsubscript{co-train} shares the same backbone model and co-training design with \QQSUMM, the lack of 
(1) opinion-level clustering of retrieved comments
and (2) limited modeling capability of LLMs 
makes this model unable to produce KPs as diverse, unique and representative as \QQSUMM.
The weak reasoning capability of LLMs for diverse opinion summarization is further exposed in the frozen Retriever + prompt LLMs setting, where LLMs even with strong modelling capability like GPT-4-Turbo struggle
to elaborate diverse and distinctive KPs from hundreds of comments.

It is worth noting that \texttt{Mistral-7B} broadly exhibits higher performance than \texttt{Vicuna-7B} across all systems based on LLM generation and in all KP quality  measurement (up to 15.32\%), largely due to its stronger modeling capability.

Frozen Retreiver + KPA baselines, despite their high performance for review summarization, is ineffective for \QQFS.
Not surprisingly PAKPA, which generates KPs based on aspect-sentiment,   broadly shows better performance than RKPA-Base, an extractive KPA system.
It is possible that multiple query-relevant opinions  
on the same aspect are expected to answer a user query, thus leading to the weak performance of PAKPA.

Our manual evaluation of KP information quality further validates the above findings, as shown by the Bradley Terry scores in Table~\ref{table:kp_quality_eval_results_T}.
Overall, \QQSUMM achieves up to 4.58 times improvements on all 7 dimensions, and are notably higher 
on \textsc{Coverage} (CV) (2.86 times), \textsc{Validity} (VL) (2.38 times), and \textsc{Sentiment} (SN) (3.5 times).

\subsubsection{KP Quantification}
Table~\ref{table:quantitative_evaluation} presents the quantification performance for different systems.
F$_1$, combining Recall and Precision, measures the overall performance of KP-comment matching for all systems. 
QuantErr (lower the better) directly measures KP quantification errors.  
Overall, \QQSUMM shows the best performance in terms of both F$_1$ (0.792 vs. 0.154) and QuantErr (4.24 vs. 30.13).  

Comparing \QQSUMM against the Retriever+LLM generation systems, namely (Retriever + LLM)\textsubscript{co-train} and Frozen Retriever + prompt LLM, we can see that, 
without clustering comments,
LLMs perform comment-KP matching and KP quantification, showing extremely low Recall (0.185--0.249), in contrast to the high Recall of \QQSUMM (0.684-0.869). 
This can be attributed to two main factors: 
(1) LLMs tend to hallucinate when generating KPs from a large set of retrieved comments, and 
(2) their limited context window restricts their ability to effectively match comments to KPs.

Comparing \QQSUMM against Retriever + KPA systems, 
our model shows up to 67.12\% improvement in quantification performance over state-of-the-art KPA system for reviews (PAKPA)~\citep{tang-etal-2024-prompted}, with a 36.53\% reduction in QuantErr.
Note that Frozen Retriever + PAKPA achieves the highest matching precision due to aspect-level opinion quantification.
However, it has low recall, possibly because it relies on aspect-based sentiment analysis of comments, which can fail to identify implicit opinions not explicitly including aspects.

As shown in Table~\ref{table:quantitative_evaluation}, results for KP-Comment Factual Alignment show that \QQSUMM and Frozen Retriever + KPA (PAKPA) achieve high factual correctness in KP generation, outperforming other systems (0.749 vs. 0.354). 
This result highlights that \QQSUMM generates KPs grounded in the retrieved comments, and similarly PAKPA generates KPs grounded in aspects.

\subsection{Ablation Study}
We evaluate the contribution of Next-KP-Generation in \QQSUMM, with results in Tables~\ref{table:ablation_single_kp_automatic_evaluation} and~\ref{table:ablation_single_kp_quantitative_evaluation} (Appendix~\ref{sec:ablation_study_results}). 
In particular, we configure a variant \QQSUMM\textsubscript{Single-KP} that replaces Next-KP-Generation with KP generation for each comment cluster.
Not including previously generated KPs as context, \QQSUMM\textsubscript{Single-KP} struggles to capture the truly representative opinion of the cluster,  
likely generating KPs with overlapping opinions, especially for comments containing multiple opinions.
Note that while its KP quality underperforms RAG baselines, its KP Quantification performance remain superior, largely attributed to KP-oriented Retrieval.

\subsection{Case studies}
We conducted case studies to evaluate the redundancy and specificity of generated KPs for a query comparing camera lenses, presented in Table~\ref{table:top_kps_compare} (Appendix~\ref{sec:kp_summary_examples}).
Overall, \QQSUMM stands out for generating KPs with minimal redundancy, high informativeness, and alignment with the query.
First, \QQSUMM reduces redundancy by effectively capturing distinct product features relevant to the user’s needs (e.g., faster aperture), 
whereas (Retriever + LLM)\textsubscript{co-train}, GPT-4-Turbo Prompt LLM, and PAKPA tend to generate repetitive and generic statements, 
such as ``The 24-70mm f/2.8 is a better lens overall.'' 
Furthermore, \QQSUMM expands feature coverage, capturing details such as Vibration Reduction (VR) technology, which several baselines fail to mention.

\subsection{Error Analysis}
Our analysis on a KP summary of \QQSUMM reports two systematic error patterns, as shown in Table~\ref{tab:qqsumm_output_example_matching}.
First, a KP can be falsely matched to comments expressing similar opinions but on different targets. 
For instance, the comment ``\emph{For a lens that is overall a rather mixed bag \dots it is very expensive.}'' was matched to KP ``\emph{The 24-120mm F4 lens has a longer zoom range and is more affordable than the 24-70mm F2.8.}''. Since the comment lacks an explicit product reference, it remains unclear whether it critiques the \emph{24-120mm F4} or the \emph{24-70mm F2.8}.
The second type of errors stems from the sentence-level quantification, where input review sentences often contain co-occurring multi-aspect opinions,
making it difficult for the Retriever to isolate distinct aspects into separate clusters.

\section{Conclusion}
\label{sec:conclusion}
In this paper, we studied a new task Quantitative Query-focused Summarization, namely \QQFS, for capturing and quantifying diverse opinions from online reviews for PQA.
We propose \QQSUMM, a few-shot summarization model based on retrieval-augmented generation where summary is generated by LLMs from groups of user opinions relevant to a query.
\QQSUMM addresses the issue of existing RAG frameworks for providing only random or major opinion in the answer. 
By extending the retriever with opinion-based clustering of relevant comments, 
our model ensures capturing more diverse and representative opinions in the summary, along with accurate quantification.
Experimental results show that our solution greatly enhances both the quality and quantitative performance of key point generation in summaries.

\section*{Acknowledgement}
This research is supported in part by the Australian Research Council Discovery Project \textbf{DP200101441}.

\section*{Limitations}
We evaluated the textual quality of generated KPs only on AmazonQ\&A, as it is the only (to our best knowledge) public dataset with abundance of online community answers written by online users usable as ground truth for our automatic evaluation.

Since we are leveraging answers from AmazonQ\&A to summarize and quantify the prevalence of query-relevant opinions from reviews regarding a query, an inevitable limitation is that key points extracted from the Q\&A answers might not fully
in line with viewpoints in reviews to answer questions.
Similarly, opinions in product reviews also may not sufficiently cover all opinions in community answers.

\section*{Ethics Statement}
We have applied ethical research standards in our organization for data collection and processing throughout our work.

The AmazonQ\&A dataset used in our experiments was publicly crowdsourced and released for the research publication for the review-based product question answering task~\citep{gupta2019amazonqa}.
The dataset was published following their ethical standard, after removing all personal information.
The answers to questions do not contain contents that are harmful to readers.

We ensured fair compensation for crowd annotators on Amazon Mechanical Turk. We setup and conducted fair payment to workers on their annotation tasks/assignments according to our organization's standards, with an estimation of the difficulty and expected time required per task based on our own experience. Especially, we also made bonus rewards to annotators who exerted high-quality annotations in their assignments.

\bibliography{anthology_0,anthology_1,custom}

\appendix
\section{Opinionated Question Classification for \AmazonKP Dataset}
\label{sec:opinionated_question_classification}
Existing online product-related questions can be categorized into two groups: 
subjective (opinionated) or objective (factal).
While subjective questions asks about positive/negative feeling or stance (e.g., whether a product is “good" or “bad"), objective questions confirms the actual product details (e.g., products properties, specific use-cases).
In E-Commerce, questions are often subjective, i.e., asking for former buyer's opinion, where 
different customers often have certain preferences over product aspects or information needs~\citep{kdd19-description,www19-tips}, leading to various expectations for the provided answers.

We extract subjective, i.e., opinionated, question from AmazonQ\&A by prompting the \texttt{Mistral-7B} open-source LLM to analyze the question and its associated answers, published by the online community. In this case, leveraging answers helps to understand the nature of the questions, thereby better reasoning whether the question is seeking for subjective information from users.
We present the few-shot prompt for classifying opinionated, i.e., subjective, questions from AmazonQ\&A in Listing~\ref{lst:few_shot_oqc}.

\begin{figure*}
    \centering
\begin{lstlisting}[caption={Few-shot prompt (2 examples) for prompting \texttt{Mistral-7B} on opinionated question classification.}, basicstyle=\small, label=lst:few_shot_oqc]
You will be provided with a question and multiple answers to that question, delimited by triple quotes.
The question was taken from a Question Answering dataset of product reviews, and can either be an opinionated or factual question.

You were tasked to classify whether the given question is an opinionated or factual question.
Factual questions ask for objective data, specifications, or information that can be definitively answered based on product facts, manual, user experience, or specifications. Factual question tends to covers unique and consistent opinions/fact in its answers.
Opinionated questions are subjective and seek insights that are based on personal use, feelings, preferences, judgments, or evaluations about a product. Opinionated question has multiple and diverse opinions in its answers.

Formally, you should perform the following steps:
1. Identify unique opinions from the answers of the given question
2. Based on the question content and the amount of opinions in the question's answer, identify the question's type. 

Note that you must briefly explain why the question is opinionated or factual before giving the final decision.

Below are some few-shot examples:

Questions: How well does it work with wireless charging
Answers: ['Unfortunately with this case installed it will not hold the phone vertically.', 'I use the case with the official wireless charger and have had no problems at all.', 'Works great. Not a fan of the dimensions.']
Type: 'Opinionated Question'

Questions: Are the shelves removeable?
Answers: ['yes, they are removeable..', 'Yes they are, you can arrange them for the size of the shot glass.']
Type: 'Factual Question'
\end{lstlisting}
\end{figure*}

\section{Qualitative Data Analysis of Opinionated Questions' Categories in \AmazonKP}
\label{sec:qual_analysis_question_amazonkp}
We further studied the utility of the \QQFS task and our by conducting qualitative data analysis to categorize possible opinionated question's type in \AmazonKP.
Based on the grounded theory methodology~\citep{charmaz2015grounded}, 
our analysis employ human-LLM collaborative annotation to iteratively code the fine-grained categories from opinionated questions.
We sampled a subset of 100 questions from \AmazonKP for data coding and intepretation.
On the subset, we start by prompting ChatGPT to identify potential categories of opinionated questions, including the categories' name and their definitions (Step 1).
Importantly, the data coding process involves human validation, in which we iteratively a human annotator iteratively evaluate the representative of generated categories while interacting with ChatGPT, and manually refine the categories where possible~\footnote{On categories requiring more fine-grained categorization, we further conduct another analysis cycle on the particular coarse-grained category, by selecting questions and answers from the specific category for analysis.} (Step 2).
Then, we prompted a \texttt{gp4-o-mini} to annotate the labels of entire questions in the subset, before asking human annotator again to validate the representative and suitability of the candidate categories on questions.
Categories with abnormal distribution, e.g., 5 times higher than others, or with high unmatching cases will be passed back to Step 2 for another iterative analysis cycles.

As a result, our analysis reported 5 categories commonly representative of question in \AmazonKP, namely, \emph{Performance}, \emph{Quality}, \emph{Recommendation}, \emph{Comparative} and \emph{Controversial}, with each the stating clearly the purpose of the users asking the questions and expected answers.
Finally, We prompted \texttt{gpt-4-o-mini} to annotate such categories on \AmazonKP's opinionated questions, and reported their taxonomy and statistics in Table~\ref{tab:amazonkp_taxonomy}.
Notably, the dominance of ``\emph{Scenario-based}'' questions underscore the importance of \QQFS for generating KP summary to answer user questions on preferences and scenarios.

\begin{table*}[!ht]
  \centering
  \small
  \scalebox{0.9}{
  \begin{tabularx}{1\textwidth}{|l|X|X|l|}
  \hline
      \textbf{Category} & \textbf{Description} & \textbf{Example} & \textbf{\# Query} \\ \hline
      Performance & Ask how well a product performs or functions in general. & How well does it work on carpet? & 376 \\ \hline
      Quality & Ask about the overall or aspect-specific quality of the product. & Is this product worth the money? & 265 \\ \hline
      Scenario-based & Ask whether a product fits specific use cases, sizes, or other products. & Does this item really stop the glare at night even in rain or snow? & 1402 \\ \hline
      Recommendation & Ask for suggestions tailored to specific issues or use cases. & What do you use to spray this stuff on your lawn? & 156 \\ \hline
      Comparative & Seeks opinions about the relative advantages or disadvantages of a product compared to others. & Would a wired keyboard/mouse be better than wireless? & 227 \\ \hline
      Controversial & Reflect dissatisfaction or complaint about a product, likely to provoke debate or controversy. & Why does this need adjustment screws?  If I have to align the laser then what’s the point? & 124 \\ \hline
  \end{tabularx}
  }
  \caption{A taxonomy of opinion questions \AmazonKP} 
  \label{tab:amazonkp_taxonomy}
\end{table*}

\section{Human Validation of GPT4's Key Point Extraction from Gold Community Answer of AmazonQ\&A}
\label{sec:kp_extraction_answer_validation}
In this experiment, we empirically validate \texttt{gpt-4-o-mini}'s performance and credibility in extracting KPs from gold community answers for AmazonKP (Stage 1 of §\ref{sec:amazonkp}).
Specifically, to maintain reasonable cost, we sampled a question, i.e., queries, from 5 common product categories of AmazonKP~\footnote{namely \emph{Home\_and\_Kitchen}, \emph{Sports\_and\_Outdoors},
\emph{Tools\_and\_Home\_Improvement}, \emph{Health\_and\_Personal\_Care} and \emph{Beauty}}, 
totaling 5 questions, 
and hired workers to annotate whether the extracted KPs matches original gold community answers of the sampled questions, which is inspired by the KP Matching evaluation of \citet{bar-haim-etal-2021-every}.
More specifically, for a given query, we asked workers to perform pairwise annotation between extracted KPs and the query's respective community answers.
While \emph{Precision} calculates the fraction of KPs matched to at least one gold answer, i.e., out of all extracted KPs how many are correctly mapped, \emph{Recall} shows the fractions of gold answers matched to at least one KP, i.e., out of all answers how many are covered by KPs.
We then macro-averaged Precision/Recall computed for every question to obtain the final values.

For human annotation, we employed 3 MTurk crowd workers on every answer-KP pair, selecting only those with an 80\% or higher approval rate and at least 10 approved tasks. 
Following \citet{bar-haim-etal-2021-every}, we exclude annotators with Annotator-$\kappa <0$ for quality control.
This score averages all pairwise Cohen's Kappa~\citep{landis1977measurement} for a given annotator, for any annotator sharing at least $50$ judgments with at least $2$ other annotators.
For labelling correct matches, we applied a strict threshold, in which 100\% votes (3 out of 3) of the annotators had to agree that the match was correct. Otherwise, it is incorrect.

\begin{table}[!ht]
  \centering
  \scalebox{0.9}{
  \begin{tabular}{|l|r|}
  \hline
      \textbf{Precision} & 87.5\%  \\ \hline
      \textbf{Recall} & 90.0\% \\ \hline
      \textbf{\# Matched Answers Per KP} & 2.39 \\ \hline
      \textbf{\# Matched KPs Per Answer} & 2.61 \\ \hline
  \end{tabular}
  }
  \caption{Performance validation of \texttt{gpt-4-o-mini}'s KP extraction from gold community answer. While precision calculates the fraction of KPs matched to at least one gold answer, recall shows the fractions of gold answers matched to at least one KP.}
  \label{table:kp_extraction_answer_validation}
\end{table}

Table~\ref{table:kp_extraction_answer_validation} presents the fraction of extracted KPs matched to at least one gold answer (Precision) and vice versa (Recall).
Overall, the experiment confirms that the extracted KPs are of high quality, 
with 90.0\% of community answers were represented with KPs (recall), while 87.5\% of the extracted KPs are verified as valid (precision).\\

Below are the match annotation guidelines for (extracted KP, gold answer) pairs:\\

In this task you are presented with a question on a product, a key point extracted from community answers answering the question, and a community answer for answering the query of that product.

You will be asked to answer the following question:"Does the key point match, i.e., represent an opinion in the community answer?"

A community answer might express opinions on multiple aspects. A key point matches a community answer if it captures the gist of the answer, or is directly supported by a point made in the community answer.

The options are:
\begin{itemize}
\item Not At All
\item Somewhat Not Well
\item Somewhat Well
\item Very Well
\end{itemize}

\section{Prompt for Key Point Extraction from Gold Community Answer of AmazonQ\&A}
\label{sec:kp_extraction_answer}
We present the few-shot prompts for extracting key points (KPs) from gold online community answers of AmazonKP in Listing~\ref{lst:few_shot_kp_extraction_answer}.

\begin{figure*}
  \centering
\begin{lstlisting}[caption={One-shot prompt (1 example) for prompting \texttt{GPT-4-o-mini} on KP Extraction from community answers.}, basicstyle=\small, label=lst:few_shot_kp_extraction_answer]
You will be provided with an opinionated question and multiple answers to that question, delimited by triple quotes.
An opinionated question seek insights of user opinions that are based on personal use, feelings, preferences, judgments, or evaluations about a product, and was taken from a Question Answering dataset of product reviews. 

You were tasked to extract a list of unique and concise key points from the list of answers to given opinionated question.
Key points are short and high quality sentences that expresses the main claims/viewpoints of users answering the opinionated question 
Note that the final extracted list of key points must directly relevant and can answer the input opinionated question.

Formally, you should perform the following steps:
1. In every answer from the list, extract all possible key point candidates.
2. From the extracted list of key point candidates, generate a list of only general and non-overlapping key points that are relevant and can answer the input opinionated question.

Below are some few-shot examples:

Questions: Can I use these for running/working out? Do they handle sweat?
Answers: ['I have seen other people using these for running/working out. These are very comfortable in your ears for long hours. As long you clean them after working out, you should be fine. These are built to last a long time.', 'I use them in the gym and on the stair climber machine.  They are fine.  Not sure about running but would think they would work ok.', "I don't know if I'll be any help, but I'll tell you about my experience nevertheless. I used it everyday in the gym & while I go for work on my bike inside my helmet. In both cases, the sweat doesn't seem to have any effect. Even during long rides, and when it rained heavily, the IE80 held up fine. The only issue you will have to worry about is the cable. Though the cables are good quality, rough usage may affect the balance in volume levels between the two channels. Though this doesn't affect the clarity, the balance can be disturbed. After a year of really rough usage, the IE80 right volume was 1-2% lower than the left [I got mine replaced for free soon after]. But, this is an issue which affects every IEM, and nothing to sweat over, since we can replace the cables if necessary. So if you don't give it a hard time, it should hold up fine.[I can't even count the times it has fallen down or swung down and taken a hit against the gym equipment, or when my phone/DAP slipped and yanked the cable]"]
Key Points: ['Comfortable for long hours', 'Built to last a long time', 'Suitable for gym and stair climber machine', 'Sweat resistant during workouts', 'Cables may be affected by rough usage']
\end{lstlisting}
\end{figure*}

\section{Annotation Details of KP Matching for \AmazonKP Dataset}
\label{sec:annotation_detail_kp_matching}
We offer \texttt{GPT-4-o-mini} with 4 options for labelling the matching status of given comment-KP pairs. 
Pairs annotated as \emph{Very Well} or \emph{Somewhat Well} by LLM then becomes \emph{candidate matching pairs}, which will be further validated by human annotation for their correctness.
For human annotation, we employed 3 MTurk crowd workers per comment-KP pair, selecting only those with an 80\% or higher approval rate and at least 10 approved tasks. 
Following \citet{bar-haim-etal-2021-every}, we exclude annotators with Annotator-$\kappa <0$ for quality control.
This score averages all pairwise Cohen's Kappa~\citep{landis1977measurement} for a given annotator, for any annotator sharing at least $50$ judgments with at least $2$ other annotators.
For labelling correct matches, at least 60\% of the annotators had to agree that the match is correct, otherwise, it is incorrect.
Comments from final matching pairs, after confirmed by human, will then be grouped by similar KPs, where the amount of matching comments per KP is the prevalence of the respective KP.\\

Below are the matching prompt for LLM and the annotation guidelines for workers validating (sentence, KP) pairs:\\

In this task, you are presented with a question on a product, a key point taken from the summary answering the question, and a sentence taken from a review of that product.

You will be asked to answer the following question: "Does the key point match, i.e, represent an opinion in the review sentence?"

A review sentence might express opinions on multiple aspects. A key point matches a sentence if it captures the gist of the sentence, or is directly supported by a point made in the sentence.

The options are:
\begin{itemize}
\item Not At All
\item Somewhat Not Well
\item Somewhat Well
\item Very Well
\end{itemize}

\section{Prompts for KP Summary Generation of \QQSUMM}
\label{sec:kpsg_prompt}
We present the instruction-finetuning prompts for KP Summary Generation of \QQSUMM in Listing~\ref{lst:few_shot_kpsg}.

\begin{figure*}
    \centering
\begin{lstlisting}[caption={Prompt for instruction-finetuning \QQSUMM's LLM for KP Summary Generation. Please refer to our released code for full prompts.}, basicstyle=\small, label=lst:few_shot_kpsg]
You will be provided with a question and a JSON list of relevant review comments, delimited by triple quotes.
The question asks the opinions of user reviews about a product, and can be answered by the list of comment clusters in the provided JSON list. Each element in the JSON has been has been clustered to represent a common opinion answering the question, accompanied by the quantity.

You were tasked to generate a quantitative summary that covers all opinions captured in the JSON list in answering the questions.

Perform the following actions to solve this task:
- For every element in the JSON list, find the key point that represent the common opinion across the comments of the cluster
- Generate a long-form quantitative summary including all extracted key points and the cluster size, following the below template:
'While answering about [Question]:
+ [Cluster size] of comments believe that [Key Point 1]
+ [Cluster size] of comments believe that [Key Point 2]
...'

Below are fundamental rules:
+ Larger cluster means higher support for the key point and with a bigger cluster size, the quantity must be higher
+ Only use number to report the cluster size for each key point, avoiding vague terms (e.g., some, most)
+ Ensure that each key point extracted from a cluster is distinctive and doesn't redundantly cover aspects mentioned in larger clusters
\end{lstlisting}
\end{figure*}

\section{Prompts for \textsc{G-Eval} Evaluation}
\label{sec:g_eval_prompt}
For implementation of \textsc{G-Eval} in our KP quality evaluation dimension (§\ref{sec:evluation_dimensions}), we specifically customize the model's original prompt for evaluating summary's \emph{relevance} and \emph{redundancy}. While the \emph{relevance} evaluation prompt is customized for evaluating sP/sF/sF1~\citep{li-etal-2023-hear} between individual generated KPs and the reference KPs, \emph{redundancy} is customized for evaluating $RD$ among generated KPs. We presented our relevance evaluation prompt in Listing~\ref{lst:g_eval_relevancy} and the redundancy evaluation prompt in Listing~\ref{lst:g_eval_redundancy}

\begin{figure*}
    \centering
\begin{lstlisting}[caption={Zero-shot prompt for \textsc{G-Eval} relevancy evaluation between generated KPs and reference KPs, supporting sP/sR/sF1 calculation.}, basicstyle=\small, label=lst:g_eval_relevancy]
You will be given one key point, short salient sentence, written to describe user opinion on a product.

Your task is to rate the summary on one metric.

Please make sure you read and understand these instructions carefully. Please keep this document open while reviewing, and refer to it as needed.

Evaluation Criteria:

Relevance (1-5) - selection of important content from the source. The summary should include only important information from the source document. Annotators were instructed to penalize summaries which contained redundancies and excess information.

Evaluation Steps:

1. Read the key point and the source key point carefully.
2. Compare the key point to the source key point and identify the main points.
3. Assess how well the key point covers the main points of the source key point, and how much irrelevant or redundant information it contains.
4. Assign a relevance score from 1 to 5.
\end{lstlisting}
\end{figure*}

\begin{figure*}
    \centering
\begin{lstlisting}[caption={Zero-shot prompt for \textsc{G-Eval} redundancy evaluation of generated KPs, supporting $RD$ calculation.}, basicstyle=\small, label=lst:g_eval_redundancy]
You will be given one key point, short salient sentence, written to describe user opinion on a product.

Your task is to rate the summary on one metric.

Please make sure you read and understand these instructions carefully. Please keep this document open while reviewing, and refer to it as needed.

Evaluation Criteria:

Redundancy (1-5) - overlapping opinion with the source. The summary should not include semantically similar opinion with the source document. Annotators were instructed to penalize summaries which contained overlapping opinion with the source.

Evaluation Steps:

1. Read the key point and the source key point carefully.
2. Compare the key point to the source key point and identify the main points.
3. Assess how much redundant opinion and information the key point covers that overlap with the source key point
4. Assign a redundancy score from 1 to 5.
\end{lstlisting}
\end{figure*}

\section{Dimensions of KP Quality Evaluation}
\label{sec:kp_quality_dimensions}
This section provides detailed descriptions of tasks and dimensions involved in our manual evaluation of the KP textual quality.
Annotators were asked to perform a pairwise comparison between two sets of KPs, each taken from a different model, generated for a specific reviewed business entity considering a specific dimension.
The annotators must answer a comparative question with respect to the evaluating dimension. (e.g., \emph{Which of the two summaries captures better \dots{}}).
For each dimension, following~\citet{friedman-etal-2021-overview}, we calculate the ranking using the Bradley-Terry model~\citep{bradley_terry}, which predicts the probability of a given participant winning a paired comparison, based on previous paired comparison results of multiple participants, and thus allows ranking them.

\begin{itemize}
    \item \textsc{Validity}: The key point in the summary should be an understandable, well-written sentence representing an opinion of the users towards the question. This would filter out sentences such as \emph{``It's rare these days to find that!''}.
    \item \textsc{Sentiment}: The key point in the summary should have a clear sentiment towards the product being questioned (either positive or negative). This would exclude sentences like \emph{``I came for a company event''}.
    \item \textsc{Informativeness}: The key point in the summary should discuss should discuss some aspects of the reviewed product and contain useful information. Any key point that is too specific or only expresses sentiment cannot be considered a good candidate.
    \item \textsc{SingleAspect}: The key point in the summary should not discuss multiple aspects (e.g., \emph{``Decent price, respectable portions, good flavor''}).
\end{itemize}

\begin{itemize}
    \item \textsc{Redundant}: Each KP should express a distinct aspect. In other words, there should be no overlap between the key points.
    \item \textsc{Coverage}: The summary, containing the set of key points, should cover a wide diversity of opinions relevant and representative to the question.
    \item \textsc{Faithfulness}: The key point in the summary should express reasonable and meaningful opinions relevant to the question raised on the product without hallucination. No conjecture or unfounded claims should arise.
\end{itemize}

\section{Pairwise KP Quality Comparison Annotation Guidelines}
\label{sec:pairwise_kp_quality_annotation_guideline}
Below are the two summaries for a product question in \emph{Tools\_and\_Home\_Improvement}, generated by two different summarization frameworks. Each summary contains several key points (i.e., salient points) generated summarizing the user opinions on different aspects. You are tasked to select which summary you think is better according to the below criteria.

\textbf{Question:} Does this tester accurately test AA Lithium? The power drop off curve is so steep. It seems unlikely...but I am hoping!.

\textbf{Criteria:} REDUNDANCY. Each key point in the summary should express a distinct aspect. In other words, there should be no overlap between the key points.

\textbf{Summary A:} ['the tester accurately tests various types of batteries, including AA Lithium, and provides accurate readings', 'there is uncertainty about the accuracy of the percentage of charge remaining for AA Lithium batteries', 'the tester does not test a specific version of AA Lithium battery (L91)', 'the tester is big and cumbersome, but effective in testing batteries under load', 'the tester requires four AA batteries to operate', 'the tester tests batteries by putting a load on them, making the readings more accurate', 'the tester tests batteries quickly, with a test taking only 3-4 seconds for a AA battery', 'the tester is expensive but worth the investment due to its accuracy and ability to save money by testing old batteries', 'the tester tests batteries of various sizes, including AA, AAA, C']

\textbf{Summary B:} ['I have compared the testers results to battery powered devices and found it does give you the true useful state of a battery. ', 'Now that I found this tester I am happy, because it tests a battery the way a battery should be tested.', 'That model also tests 6v lithium 2CR5 used in some older cameras, which the current tester does not since the times have moved on.']

The options are:
\begin{itemize}
\item Summary A
\item Summary B
\end{itemize}

\section{GPT4's Comment-KP Matching Annotation against Human Judgement}
\label{sec:gpt4_annotation_validate}
To validate \texttt{gpt-4-o-mini}'s annotation performance and credibility, we conduct an experiment to measure LLM annotation judgement, as utilized for the KP-comment matching evaluation in our main experiment, in agreement with human (gold) preference.
We sampled a subset of 5 queries from the test set in our main experiment and hired workers to annotate the correctness of comment-KP pairs produced as the results of our framework's quantification outcome.
Note that these sampled pairs are part of the our main test set and have already been annotated for LLM's labels in our main experiment.
For human annotation, we employed 6 MTurk crowd workers on every comment-KP pair, selecting only those with an 80\% or higher approval rate and at least 10 approved tasks. 
Following \citet{bar-haim-etal-2021-every}, we exclude annotators with Annotator-$\kappa <0$ for quality control.
This score averages all pairwise Cohen's Kappa~\citep{landis1977measurement} for a given annotator, for any annotator sharing at least $50$ judgments with at least $5$ other annotators.
For labelling correct matches, at least 60\% of the annotators had to agree that the match is correct, otherwise, it is incorrect.
In this experiment, we measured the accuracy, and conducted a Pearson correlation ($r$) test of \texttt{gpt-4-o-mini}'s annotation performance against human judgement, with results reported in Table~\ref{table:gpt4_annotation_validate}. For $r$ test, we set the null hypothesis as \texttt{gpt-4-o-mini}'s and Mturk annotated labels are independent.

From Table~\ref{table:gpt4_annotation_validate}, we saw signficant small p-value, which indicates strong evidence against the null hypothesis.
Importantly, we also recorded Spearman's rank correlation coefficient to be relatively closed to 1. This implies that there is a statistically significant positive correlation between \texttt{gpt-4-o-mini} and Mturk annotated labels, which substantiates our decision of using \texttt{gpt-4-o-mini} for comment-KP matching evaluation.

\begin{table}[!ht]
  \centering
  \scalebox{0.9}{
  \begin{tabularx}{0.5\textwidth}{|l|X|}
  \hline
      \textbf{Pearson correlation ($r$)} & 0.647 \\ \hline
      \textbf{p\_value} & 5.342e-16\\ \hline
      \textbf{Accuracy} & 0.807 \\ \hline
  \end{tabularx}
  }
  \caption{Performance valiation of GPT4's comment-KP matching annotation against human judgement}
  \label{table:gpt4_annotation_validate}
\end{table}

Below are the match annotation guidelines for (sentence, KP) pairs:\\

In this task, you are presented with a question on a product, a key point taken from the summary answering the question, and a sentence taken from a review of that product.

You will be asked to answer the following question: "Does the key point match, i.e, represent an opinion in the review sentence?"

A review sentence might express opinions on multiple aspects. A key point matches a sentence if it captures the gist of the sentence, or is directly supported by a point made in the sentence.

The options are:
\begin{itemize}
\item Not At All
\item Somewhat Not Well
\item Somewhat Well
\item Very Well
\end{itemize}

\section{Clustering Algorithm of KP-Oriented Retrieval in \QQSUMM}
\label{sec:ablation_clustering_algorithm_eval}
To validate other clustering techniques, we have developed an additional baseline that employs either HDBSCAN~\citep{mcinnes2017hdbscan} or K-Means clustering algorithm for grouping similar comments by the Retriever, following our main experimental setup and configuration in Section~\ref{sec:kp_quantification}.
Better than K-Means, HDBSCAN can automatically detect the number of clusters without pre-defined parameters and is used in a previous KPA work~\citep{li-etal-2023-hear}. We compare the factual alignment of KP-comment pairs (measured by AlignScore) across clustering methods in Table~\ref{table:ablation_clustering_quantitative_evaluation}, using our best model configuration (Contriever + Mistral):

While both HDBSCAN and K-Means perform reasonably, they are consistently outperformed by our specialized clustering approach. More specifically, although HDBSCAN or K-Means achieves relatively comparable matching Precision with our clustering algorithm, our algorithm can capture comments more sufficiently (much higher Recall) than HDBSCAN and K-Means. This is mostly because our algorithm contains more tuneable clustering parameters and operations that are specifically optimized for the QQSUM problem.

\begin{table*}[t]
  \centering
  \scalebox{0.65}{
  \begin{tabular}{lccccc}
  \toprule
  {} & \multicolumn{4}{c}{KP-Comment Matching} & \multicolumn{1}{c}{\makecell{KP-Comment Factual Alignment}} \\
  \cmidrule(r){2-5} \cmidrule(r){6-6}
  & P & R & F1 & QuantErr$\downarrow$ & AlignScore \\
  \midrule
  Our proposed clustering algorithm	& \textbf{0.694} & \textbf{0.869} & \textbf{0.792} & \textbf{04.24} & \textbf{0.749} \\
  HDBSCAN clustering algorithm	& 0.682 & 0.507 & 0.582 & 11.47 & 0.718 \\
  K-Means clustering algorithm (n\_clusters = 3)	& 0.677 & 0.424 & 0.522 & 15.50 & 0.681 \\
  \bottomrule
  \end{tabular}
  }
  \vspace{-3mm}
  \caption{\label{table:ablation_clustering_quantitative_evaluation}
  KP-Comment matching performance and factual consistency of generated summary between different clustering methods applied for KP-oriented Retrieval of \QQSUMM. The experiment was conducted with the \texttt{Mistral} configuration for \QQSUMM, proven to have superior performance than \texttt{Vicuna} from Table~\ref{table:automatic_evaluation}.
  }
\end{table*}

\section{Ablation Study: Single-KP Generation vs KP Summary Generation in \QQSUMM}
\label{sec:ablation_kp_summary_generation_eval}
We conducted an ablation study to evaluate the impact of KP Summary Generation on \QQSUMM, with KP quality and KP-comment matching and factual consistency performance presented in Table~\ref{table:ablation_single_kp_automatic_evaluation} and~\ref{table:ablation_single_kp_quantitative_evaluation} respectively.
To this end, we configure \QQSUMM\textsubscript{Single-KP}, a variant that generates one KP at a time for each comment cluster formed by KP-oriented Retrieval.

Overall, not including previously generated KPs as context, \QQSUMM\textsubscript{Single-KP} 
struggles to capture the truly representative opinion of the cluster,  
likely generating KPs with overlapping opinions, especially for comments containing multiple opinions.

\begin{table*}[t]
  \centering
  \scalebox{0.6}{
  \begin{tabular}{lcccccccccccccccccc}
  \toprule
  {} & \multicolumn{3}{c}{ROUGE} & \multicolumn{5}{c}{BERTScore} & \multicolumn{5}{c}{BLEURT} & \multicolumn{5}{c}{G-Eval-4} \\
  \cmidrule(r){2-4} \cmidrule(r){5-9} \cmidrule(r){10-14} \cmidrule(r){15-19}
   & R-1 & R-2 & R-L & sP & sR & sF1 & RD$\downarrow$ & Rel & sP & sR & sF1 & RD$\downarrow$ & Rel & sP & sR & sF1 & RD$\downarrow$ & Rel \\
  \midrule
  \rowcolor{lightgray}
  \multicolumn{19}{l}{$\mathbf{\QQSUMM}$ (Ours)}\\
  \specialrule{0em}{1pt}{1pt}
  \hspace{3mm} + Mistral & \textbf{0.256} & 0.061 & \textbf{0.220} & \textbf{0.39} & \textbf{0.29} & \textbf{0.33} & \textbf{0.37} & \textbf{0.27} & \textbf{0.51} & \textbf{0.41} & \textbf{0.46} & \textbf{0.49} & \textbf{0.45} & \textbf{4.52} & \textbf{4.29} & \textbf{4.40} & \textbf{2.43} & \textbf{4.05} \\ %
  \hspace{3mm} + Vicuna & 0.222 & \textbf{0.078} & 0.204 & 0.38 & 0.26 & 0.31 & 0.53 & 0.25 & 0.49 & 0.39 & 0.44 & 0.54 & 0.41 & 4.47 & 4.25 & 4.36 & 2.45 & 3.68 \\ 
  \rowcolor{lightgray}
  \multicolumn{19}{l}{\QQSUMM\textsubscript{Single-KP}}\\  
  \specialrule{0em}{1pt}{1pt}
  \hspace{3mm} + Mistral & 0.191 & 0.035 & 0.160 & 0.29 & 0.22 & 0.25 & 0.48 & 0.22 & 0.48 & 0.39 & 0.43 & 0.62 & 0.39 & 4.21 & 4.22 & 4.22 & 2.51 & 3.14 \\
  \hspace{3mm} + Vicuna & 0.171 & 0.045 & 0.154 & 0.22 & 0.17 & 0.19 & 0.57 & 0.20 & 0.48 & 0.38 & 0.42 & 0.66 & 0.36 & 4.10 & 4.12 & 4.11 & 2.60 & 2.87 \\
  \bottomrule
  \end{tabular}
  }
  \caption{\label{table:ablation_single_kp_automatic_evaluation}
  KP-level textual quality evaluation of generated summary between full implementation of \QQSUMM and without (w/o) KP Summary Generation.
  sP, sR and sF1 refer to Soft-Precision, Soft-Recall, and Soft-F1 respectively based on set-level evaluation method against 
  reference KPs in gold answer.
  \textsc{G-Eval}-4 asks GPT-4 to score a summary from 1-5.
  }
\end{table*}

\label{sec:ablation_study_results}
\begin{table*}[t]
  \centering
  \scalebox{0.6}{
  \begin{tabular}{lcccccc}
  \toprule
  {} & \multicolumn{4}{c}{KP-Comment Matching} & \multicolumn{2}{c}{\makecell{KP-Comment Factual Consistency}} \\
  \cmidrule(r){2-5} \cmidrule(r){6-7}
  & P & R & F1 & QuantErr$\downarrow$ & \makecell{AlignScore \\ (cluster-level)} & \makecell{AlignScore \\ (retrieval-level)} \\
  \midrule
  \rowcolor{lightgray}
  \multicolumn{7}{l}{$\mathbf{\QQSUMM}$ (Ours)}\\
  \specialrule{0em}{1pt}{1pt}
  \hspace{3mm} + Mistral & 0.694 & \textbf{0.869} & \textbf{0.792} & \textbf{04.24} & \textbf{0.749} & \textbf{0.826} \\
  \hspace{3mm} + Vicuna & 0.538 & 0.684 & 0.602 & 07.83 & 0.630 & 0.690 \\
  \rowcolor{lightgray}
  \multicolumn{7}{l}{\QQSUMM\textsubscript{Single-KP}}\\  
  \specialrule{0em}{1pt}{1pt}
  \hspace{3mm} + Mistral & 0.640 & 0.520 & 0.574 & 17.84 & 0.682 & 0.741 \\
  \hspace{3mm} + Vicuna & 0.598 & 0.471 & 0.527 & 22.63 & 0.601 & 0.660 \\
  \bottomrule
  \end{tabular}
  }
  \caption{\label{table:ablation_single_kp_quantitative_evaluation}
  KP-Comment matching performance and factual consistency of generated summary between full implementation of \QQSUMM and without (w/o) KP Summary Generation.
  }
\end{table*}

\section{Example output of \QQSUMM and Baselines}
\label{sec:kp_summary_examples}
We report the example output of query-relevant comment clusters and KP summary produced by \QQSUMM in Table~\ref{tab:qqsumm_example} and~\ref{tab:qqsumm_output_example_matching}, and further compare top 5 key points, extracted from the summary of \QQSUMM and the baselines in Table~\ref{table:top_kps_compare}.
Overall, \QQSUMM stands out for generating KPs with minimal redundancy, higher informativeness, and better alignment with the query.

\begin{table*}[t!]
  \scriptsize
  \begin{tabularx}{\linewidth}{m{1.5cm}|X}
      \toprule
      \textbf{Query} & How does this \emph{Nikon 24-120mm F4} lens compared with the \emph{24-70mm F2.8} as a general walk around lense? \\
      \midrule
      \textbf{Query-Relevant Comment Clusters} & 
      \ctext[RGB]{186,248,255}{Cluster1:} \newline
      \tabitem I like the 24-70 better but \emph{this lens is a good all around and compact optic for everyday shooting.} \newline
      \tabitem As has been said many times before: "the best lens is the one you will use", and I know \emph{I wouldn not use the 24-70mm F2.8 because it\'s too heavy and bulky to take on backpacking/camping trips and when traveling abroad.}\newline
      \tabitem This is \emph{the one lens which could replace 24-70 / 2.8,  70-200 2.8 VR II (up to some extent) for "everyday" use.'} \newline
      \tabitem \dots \newline
      \ctext[RGB]{218,224,236}{Cluster2:} \newline
      \tabitem I have an upcoming stay in Spain, and \emph{I'm seriously considering taking this lens instead of my AF-S 24-70 because of its size and zoom range.} \newline
      \tabitem My only complaint is the price tag: for a lens that is overall a rather mixed bag (depending on what you're looking for you might be very happy with it, or very disappointed) \emph{it is very expensive}. \newline
      \tabitem \emph{The 24-120 has good reach, good image quality, not heavy, not that expensive for what it can do} (constant f/4 in a zoom is very respectable) and it's also the only usable medium-telephoto FX zoom from Nikon with the VR technology. \newline
      \tabitem \emph{For a 5x zoom to be able to compete with a 3x zoom costing over \$500 more(the Nikkor 24-70mm F2.8) should only mean that the 5x zoom is a remarkable lens.}\newline
      \tabitem \dots \newline
      \ctext[RGB]{238,206,206}{Cluster3:} \newline
      \tabitem \emph{For one thing, 24 70 is know to have better quality than this one.} \newline
      \tabitem \emph{The range from 70 to 120 is not as important as a better overall quality.} \newline
      \tabitem \emph{This is probably not the best lens to use for portraits because it's just not fast enough (f-stop)}, but for travel, chasing you kids around, or any other every day shooting this lens is perfect. \newline
      \tabitem \emph{The biggest pro for the 24-70mm is the extra 1 stop of light, slightly quicker autofocus speed, and of course the corresponding softer bokeh due to the 1 stop aperture opening.}\newline
      \tabitem \dots \newline
      \\
      \midrule
      \midrule
      \textbf{KP Summary} & 
      While comparing the Nikon 24-120mm F4 lens with the 24-70mm F2.8 lens as a general walk-around lens:\newline
      + 135 of comments believe that \ctext[RGB]{186,248,255}{the Nikon 24-120mm F4 lens is relatively lightweight and compact, making it easy to carry around and use for extended periods of time.}\newline
      + 11 of comments suggest that \ctext[RGB]{218,224,236}{the 24-120mm F4 lens has a longer zoom range and is more affordable than the 24-70mm F2.8.}\newline
      + 9 of comments \ctext[RGB]{238,206,206}{prefer the 24-70mm F2.8 for its better image quality and faster aperture.} \newline
      \dots
      \\
      \midrule
      \bottomrule
  \end{tabularx}
  \caption{Example output of query-relevant comment clusters and KP summary produced by \QQSUMM, given a query, i.e., question, from AmazonQ\&A.
  Comment clusters to a particular KP are marked in the same color as the corresponding bullet in the summary.
  The \textit{relevant opinion} in each comment that directly support the corresponding KP is \textit{italicized}.
  }
  \label{tab:qqsumm_example}
\end{table*}

\begin{table*}
    \begin{center}
        \begin{small}
            \begin{tabularx}{1\textwidth}{|p{0.3\textwidth}|l|p{0.5\textwidth}|}
                \hline
                \multicolumn{3}{|X|}
                {
                    \textbf{Query}: How does this \emph{Nikon 24-120mm F4} lens compared with the \emph{24-70mm F2.8} as a general walk around lense?
                }\\ \hline
                \textbf{Key Point}&\multicolumn{1}{c|}{\textbf{Prevalence}}&\textbf{Matching Comments}\\
                \hline
                \hline
                The Nikon 24-120mm F4 lens is relatively lightweight and compact, making it easy to carry around and use for everyday shooting.& 135 & I like the 24-70 better but \emph{this lens is a good all around and compact optic for everyday shooting.} \\
                \cline{3-3}
                && As has been said many times before: "the best lens is the one you will use", and I know \emph{I wouldn not use the 24-70mm F2.8 because it\'s too heavy and bulky to take on backpacking/camping trips and when traveling abroad.} \\
                \hline
               The 24-120mm F4 lens has a longer zoom range and is more affordable than the 24-70mm F2.8. & 11 & I have an upcoming stay in Spain, and \emph{I'm seriously considering taking this lens instead of my AF-S 24-70 because of its size and zoom range.} \\
                \cline{3-3}
                && My only complaint is the price tag: for a lens that is overall a rather mixed bag (depending on what you're looking for you might be very happy with it, or very disappointed) \emph{it is very expensive}. \\
                \hline
                Prefer the 24-70mm F2.8 for its better image quality, faster aperture and better for wide shot.& 9 & \emph{For one thing, 24 70 is know to have better quality than this one.} \\
                \cline{3-3}
                && \emph{The range from 70 to 120 is not as important as a better overall quality.} \\
                \hline
            \end{tabularx}
            \caption{Top 3 key points mentioned in the KP summary produced by \QQSUMM for  answering a query from \AmazonKP. For each key point, we show the prevalence, i.e., number of matching comments (with similar aspects of the same cluster), 
            and two top matching comments. The \textit{relevant opinion} in each comment that directly support the corresponding KP is \textit{italicized}. \label{tab:qqsumm_output_example_matching}} 
        \end{small}
    \end{center}
\end{table*}

\begin{table*}[h!]
  \centering
  \small
  \aboverulesep = 0pt
  \belowrulesep = 0pt
  \begin{tabularx}{\textwidth}{|X|X|X|X|X|}
  \hline
      \multicolumn{5}
      {|>{\hsize=\dimexpr5.5\hsize+2\tabcolsep+\arrayrulewidth\relax}X|}
      {
        \textbf{Query}: How does this \emph{Nikon 24-120mm F4} lens compared with the \emph{24-70mm F2.8} as a general walk around lense?
      }\\ \hline
      \multicolumn{1}{|X}{\textbf{\QQSUMM}} & \multicolumn{1}{|X}{\textbf{(Retriever+\newline LLM)\textsubscript{co-trained}}} & \multicolumn{1}{|X}{\textbf{Contriever + \newline GPT-4-Turbo}} & \multicolumn{1}{|X}{\textbf{Contriever + \newline PAKPA}} & \multicolumn{1}{|X|}{\textbf{Contriever + \newline RKPA-Base}} \\ \hline
      The Nikon 24-120mm F4 lens is relatively lightweight and compact, making it easy to carry around and use for everyday shooting & The 24-120mm f/4 offers more reach and versatility than the 24-70mm f/2.8. & The 24-120mm lens offers good versatility and value for general use & The 24-120 lens is preferred over the Nikkor 24-70mm F2.8. due to its lighter weight. & The 24-120 is finally at a stage where you can carry it around on your FX camera and have no regrets. \\ \hline
      The 24-120mm F4 lens has a longer zoom range and is more affordable than the 24-70mm F2.8. & \ctext[RGB]{255,0,0}{The 24-120mm f/4 is lighter} and more affordable than the 24-70mm f/2.8. & The 24-70mm lens has superior \ctext[RGB]{255,0,0}{image quality and performance} & \hl{Best 4+ star walk-around lens on the market.} & If you want a 4+ star walk-around lens that covers a \colorbox{red}{great range}, this is the best on the market. \\ \hline
      Prefer the 24-70mm F2.8 for its better image quality, faster aperture and better for wide shot. & \hl{The 24-70mm f/2.8 is a better lens overall.} & the 24-70mm lens is preferred for its \ctext[RGB]{255,0,0}{optical superiority}. & The 24-70mm lens is highly recommended for wide shots. & The 24-70mm lens is more expensive but buy it if you need to shoot wide. \\ \hline
      The 24-120mm F4 lens has good image quality, with sharpness and contrast that is comparable to the 24-70mm f/2.8 & \ctext[RGB]{255,0,0}{The 24-120mm f/4 is too heavy.} & \hl{The 24-120mm lens is a more practical choice for everyday use.} & The Nikon 24-120 lens has good contrast compared to the Nikon 24-70 lens. & I briefly considered the 24-70, but the \colorbox{red}{extra reach}, vibration reduction, and lower price point sold me on this lens. \\ \hline
      The 24-120mm F4 lens has good Vibration Reduction (VR) technology that helps to reduce camera shake when taking handheld shots. & The 24-120mm f/4 has image stabilization, which is a significant advantage for handheld shots. & The 24-120mm f/4 has image stabilization for handheld shots. & N/A & N/A \\ \hline
      \multicolumn{5}{|c|}{\dots} \\\hline
  \end{tabularx}
  \caption{\label{table:top_kps_compare}
  Top 5 key points, extracted from the summary of \QQSUMM and the baselines, ranked by their prevalence on an example query from \AmazonKP. Overlapping opinions across KPs are highlighted \colorbox{red}{red}. KPs lacking of informativeness are highlighted \colorbox{yellow}{yellow}
  }
\end{table*}

\end{document}